%% file: sn-article.tex
\documentclass[pdflatex,sn-mathphys-num,iicol]{sn-jnl}

\usepackage{graphicx}%
\usepackage{multirow}%
\usepackage{amsmath,amssymb,amsfonts}%
\usepackage{amsthm}%
\usepackage{mathrsfs}%
\usepackage[title]{appendix}%
\usepackage{xcolor}%
\usepackage{textcomp}%
\usepackage{manyfoot}%
\usepackage{booktabs}%
\usepackage{algorithm}%
\usepackage{algorithmicx}%
\usepackage{algpseudocode}%
\usepackage{listings}%
\usepackage[normalem]{ulem}
\usepackage{subcaption}

\begin{document} 

\title[Article Title]{HSTR-Net: Reference Based Video Super-resolution  with Dual Cameras}

\author*[1]{\fnm{H. Umut} \sur{Suluhan}}\email{suluhan@arizona.edu}
\author[2]{\fnm{Abdullah Enes} \sur{Doruk}}\email{enes.doruk@ozu.edu.tr}
\author[2]{\fnm{Hasan F.} \sur{Ates}}\email{hasan.ates@ozyegin.edu.tr}
\author[3]{\fnm{Bahadir K.}\sur{Gunturk}}\email{bkgunturk@medipol.edu.tr}

\affil*[1]{\orgdiv{Electrical and Computer Engineering Department}, \orgname{University of Arizona}, \orgaddress{\city{Tucson}, \postcode{85721}, \state{AZ}, \country{USA}}}

\affil[2]{\orgdiv{Faculty of Engineering}, \orgname{Ozyegin University}, \orgaddress{\city{Istanbul}, \postcode{34794}, \country{Turkey}}}

\affil[3]{\orgdiv{School of Engineering and Natural Sciences}, \orgname{Istanbul Medipol University}, \orgaddress{\city{Istanbul}, \postcode{ 34810}, \country{Turkey}}}

\abstract{ High-spatio-temporal resolution (HSTR) video recording plays a crucial role in enhancing various imagery tasks that require fine-detailed information. State-of-the-art cameras provide this required high frame-rate and high spatial resolution together, albeit at a high cost. To alleviate this issue, this paper proposes a dual camera system for the generation of HSTR video using reference-based super-resolution (RefSR). One camera captures high spatial resolution low frame rate (HSLF) video while the other captures low spatial resolution high frame rate (LSHF) video simultaneously for the same scene. A novel deep learning architecture is proposed to fuse HSLF and LSHF video feeds and synthesize HSTR video frames. The proposed model combines optical flow estimation and (channel-wise and spatial) attention mechanisms to capture the fine motion and complex dependencies between frames of the two video feeds. Simulations show that the proposed model provides significant improvement over existing reference-based SR techniques in terms of PSNR and SSIM metrics. The method also exhibits sufficient frames per second (FPS) for aerial monitoring when deployed on a power-constrained drone equipped with dual cameras. 
 }

\maketitle

\input{tex/10_introduction}

\input{tex/20_related_work}

\input{tex/30_method}

\input{tex/40_dataset}

\input{tex/50_results}

\input{tex/60_ablation}

\input{tex/70_conclusion.tex}

\bibliography{refs}
\end{document}

%% file: tex/10_introduction.tex
\section{Introduction}


High spatio-temporal resolution (HSTR) cameras provide detailed information in both spatial and temporal dimensions, benefiting many video-processing tasks. While these cameras record at high frame rates, commercial models often compromise on spatial quality~\citep{cheng2020dual}. For instance, some cellphone cameras can record slow-motion videos at 1080p with a frame rate of 120 or 240 fps, but this drops to 60 fps when shooting in 4K.  Professional high-speed cameras capture recordings with high spatial and temporal quality, but they come at a high retailing price~\citep{cheng2020dual}. Also, ultra-wide angle cameras can provide high-resolution outputs of wide-area scenes, albeit at a lower frame rate. To address these limitations and achieve HSTR video feeds, video super-resolution has emerged as a promising approach that produces high-quality video from one or more low-resolution feeds of the same scene.

Video super-resolution can be leveraged to produce high-quality video feeds from dual camera sources, suitable for a variety of scenarios including wide-area monitoring, mobile phone video generation, etc. Wide-area monitoring over large regions necessitates the identification and monitoring of numerous relatively small and distant objects. It is employed for the purpose of identifying vehicles, pedestrians, anomalies, and security-related occurrences across extensive regions, utilizing either aerial vehicles flying at high altitudes or fixed camera poles. Nevertheless, as the spatial resolution of these objects decrease, the task of precise detection becomes increasingly challenging, relying more heavily on the object's motion characteristics. Consequently, the camera's frame rate becomes a critical factor in achieving accurate motion estimation and analyzing complex dynamic scenarios. Therefore, wide-area monitoring demands wide-angle, high-resolution cameras capable of continuously monitoring expansive areas, preferably at a high frame rate. However, wide-angle, high-resolution cameras and those with high frame rates are both costly and bulky for deployment on small scale drones. Another application scenario is for mobile phones with multiple cameras featuring both wide-angle (low-detail) and telephoto (high-detail) lenses. The widespread use of multiple cameras mounted on resource-constrained devices aligns well with the concept of  generating a high-quality high frame rate wide-angle video by utilizing the telephoto camera feed  as HR reference and wide-angle camera feed as LR frame.  

Comprehensive studies have been conducted to improve the quality of video footage captured by commercial cameras. The pursuit of enhanced resolution has been investigated in two main aspects: spatial super-resolution (SR), which involves synthesizing high-resolution ($HR$) frames from corresponding low-resolution ($LR$) frames, and temporal frame interpolation (TFI), aimed at enhancing the temporal resolution of videos with low frame rates. However, the challenge lies in estimating the missing spatial and temporal information for these tasks. TFI tends to introduce motion blur and artifacts due to its assumption of linear motion between adjacent frames, while spatial interpolation often results in overly smoothed frames, thereby missing essential high-frequency components~\citep{xue2019video}. Single image super-resolution (SISR) methods~\citep{dong2015image, kim2016accurate, lim2017enhanced, ledig2017photo, denton2015deep, wang2018esrgan, zhang2012single, freeman2002example} also often yields blurred results due to its inability to accurately recover high-frequency details that are absent in the $LR$ image~\citep{zhang2019image, shim2020robust, yang2020learning, lu2021masa, lin2021residual, huang2022task, liu2022dsma}.  

An alternative approach involves the application of computational imaging techniques, which leverage multiple cameras with varying resolutions and frame rates to produce high spatio-temporal resolution (HSTR) videos~\citep{cheng2020dual, zheng2018crossnet}. The dual camera systems offer cost-effectiveness and adaptability to various scenarios. We adopt a similar strategy and put forth the idea of merging two camera feeds: one offering high spatial resolution but low frame rate (HSLF), and the other providing low spatial resolution but high frame rate (LSHF). This fusion enables generation of a high spatio-temporal resolution (HSTR) video, addressing the limitations associated with single-camera resolution enhancement methods. To realize the dual camera setup, we follow the reference-based super-resolution (RefSR) approach that utilizes a high-resolution reference ($REF$) image to  perform texture transfer between high-quality spatial components and low-quality counterparts in the $LR$ image. RefSR consistently produces quantitatively and visually better results when compared to SISR, mainly due to the effective transfer of relevant textures from the $REF$ image to the $LR$ image~\citep{cheng2020dual,  zhang2019image, shim2020robust,  yang2020learning, lu2021masa, lin2021residual,huang2022task, zheng2018crossnet,jiang2021robust,  wang2021dual, xia2022coarse, lee2022reference}.

RefSR-based networks typically operate in the spatial domain, employing a correspondence matching strategy between all possible pairs of $REF$ and $LR$ patches for subsequent texture transfer. However, exhaustive patch-matching strategy struggles to establish pixel-level correspondences in the temporal domain, a challenge frequently encountered in processing multiple video feeds, despite the high computational demand. 
In the realm of RefSR, attention-based mechanisms play a crucial role in determining the correspondences between patches in $LR$ and $REF$ images~\citep{yang2020learning, lu2021masa, lin2021residual, liu2022dsma}. These mechanisms encompass both hard and soft-attention techniques, facilitating the matching of similar patches while assigning weights to these correspondences to balance their influence based on patch similarity. Despite its high-quality outcome, this approach requires searching and matching of all possible patch pairs, which results in quadratic computational complexity~\citep{xia2022coarse}. Hence, there is an inevitable need for an efficient method capable of optimizing both memory usage and execution time, allowing to minimize the computational complexity and yet exploit the features generated by attention mechanisms to their fullest extent.

Considering the aforementioned studies, there is a clear need for RefSR networks that are reliable, low in complexity and highly accurate, while also accommodating the energy and memory constraints of embedded systems. In this work, we present a configuration involving a camera recording high spatial resolution (HSR) - low frame rate (LFR) video feed, alongside a secondary camera capturing low spatial resolution (LSR) - high frame rate (HFR) video feed. Our HSTR-Net model combines these two video feeds to generate high spatial resolution high frame rate (HSTR) video content. HSTR-Net is a deep neural network designed to be lightweight and capable of almost real-time processing. It leverages both a reference ($REF$) frame and a low-resolution ($LR$) frame inspired by reference-based super-resolution techniques. This approach enables HSTR-Net to synthesize missing frames in the HSR-LFR video using the corresponding $LR$ frame from the LSR-HFR video feed. To achieve this, HSTR-Net employs motion estimation/compensation and also incorporates an attention-based patch-matching mechanism. Subsequently, a fusion process is applied to generate the missing $HR$ frame. A distinctive feature of our approach is the novel patch-matching strategy that transforms the self-attention mechanism from the Swin transformer~\citep{liu2021swin} into a patch correspondence solution for RefSR. This innovative approach allows for high processing speeds on embedded systems, as it conducts patch matching through a lightweight attention mechanism, as opposed to previous RefSR methods that involve comparing all possible pairs of patches from the two input frames. 

HSTR-Net deviates from our previous work~\citep{suluhan2022dual} in terms of both complexity and quantitative performance.~\citep{suluhan2022dual} requires a total of $5$ input frames to construct the corresponding high-resolution frame, which in turn demands increased computational power, thereby resulting in a higher energy consumption footprint. In addition, the model in ~\citep{suluhan2022dual} is only applicable in scenarios where the $LR$ and $REF$ frames are temporally close to each other, thereby limiting its applicability for high frame rate increases. On the other hand, HSTR-Net offers a more generalized solution which can be successfully utilized for low frame rate video sequences with fast and complex motion, even when it is hard to achieve a precise registration of features between the $LR$ and $REF$ frames.

HSTR-Net stands out as a lightweight RefSR network that demonstrates competitive performance both quantitatively and qualitatively, making it suitable for deployment on embedded systems. HSTR-Net outperforms state-of-the-art in RefSR by $0.28$ dB on the Vimeo90K\citep{xue2019video} septuplet dataset while significantly reducing processing time. When evaluated on the higher-resolution  VisDrone\citep{VisDrone}  aerial dataset, HSTR-Net excels with a remarkable $7.14$ dB improvement and an average run-time reduction of $\%54$ over the next fastest RefSR model. Furthermore, HSTR-Net's superior performance is also assessed on the MAMI\citep{MAMI} dataset, a wide-area surveillance (WAS) dataset recorded using a multi-camera setup. 
Benchmark results on the MAMI dataset underscore HSTR-Net's adaptability and generalization capability, showcasing its ability to outperform prior approaches by 1.1 to 4.76 dB while maintaining high visual quality. Additionally, HSTR-Net is deployed on Nvidia Jetson AGX Xavier Developer Kit~\citep{xavier}, achieving a processing rate of $4.62$ frames per second.

The novel contributions of this paper can be summarized as follows:
\begin{itemize}
    \item HSTR-Net is a lightweight deep learning-based RefSR model that  can provide  high-speed processing on resource-constrained embedded platforms. 
    \item HSTR-Net architecture introduces novel and low-complexity  patch correspondence using cross-frame attention based on a modified Swin transformer.
    \item  Optical flow-based warped features and patch-based attention features between the $REF$ and $LR$ frames are fused in HSTR-Net architecture for optimal high-frequency detail and texture transfer.  
    \item HSTR-Net achieves state-of-the-art results, both quantitatively and visually, for Vimeo, VisDrone and MAMI dual-camera datasets.
\end{itemize}

Section \ref{related} discusses the previous work in SR, VFI, and RefSR. Section \ref{Methodology}  gives the architectural details of HSTR-Net. Section \ref{sims}  provides details of training, simulations, and evaluation of results. Section \ref{conclusion}  concludes the paper with a discussion of future work.

%% file: tex/20_related_work.tex
\section{Related Work}
\label{related}
\subsection{Reference-Based Super Resolution (RefSR)}

Single Image Super-Resolution (SISR) encounters challenges in incorporating high-frequency details into the super-resolution process, primarily because it relies solely on $LR$ images. In response to this limitation, Reference-based Super-Resolution (RefSR) is an innovative approach that introduces an additional image or frame as reference. This reference frame serves as a means to infuse high-frequency details, complementing the low-resolution features and enhancing texture transfer and fusion within the super-resolution process.

The literature presents various RefSR methods that rely on patch matching between the reference and low-resolution images, followed by the transfer of textures from the $REF$ to the $LR$ image~\citep{zhang2019image, yang2020learning, lu2021masa, lin2021residual, jiang2021robust,  wang2021dual, xia2022coarse, lee2022reference}. Despite achieving competitive results, these methods face challenges in capturing fine-grained information between successive frames in a video feed. Other studies adopt an approach involving optical flow estimation and compensation on neighboring frames, followed by a fusion process to obtain the final result~\citep{cheng2020dual, zheng2018crossnet}. However, optical flow alone struggles to handle the rapid movement and frequent high rotation characteristics often associated with drones.

~\citep{zhang2019image, yang2020learning} rely on exhaustive patch matching between the $REF$ and $LR$ images, which can be computationally intensive and thereby unsuitable for deployment on resource-constrained platforms.~\citep{jiang2021robust} propose two separate methodologies, including a correspondence network that learns robust transformation matching for addressing scale and rotation misalignments, and a teacher-student correlation approach that incorporates distilled knowledge for texture transfer. Nevertheless, a performance degradation occurs when dealing with substantial scale misalignments between the $REF$ and $LR$ images.


~\citep{lu2021masa, xia2022coarse} offer an alternative approach to the problem by introducing coarse-to-fine correspondence matching instead of exhaustively iterating through all potential matches, which reduces the model complexity while achieving satisfying results.~\citep{huang2022task} divides the RefSR task into two distinct stages. Initially, they apply single image super-resolution (SISR) and then leverage the generated SR image to facilitate texture transfer. This division helps address the issues of reference-underuse and reference-misuse, which pertain to insufficient and irrelevant texture transfers, respectively.~\citep{guodouble} introduces a dual-search strategy, initially filtering out irrelevant textures from the $REF$ data during the first pass to narrow down the search space. In the subsequent stage, refined search space allows for accurate texture transfer between the  $REF$ and $LR$ images.
~\citep{mei2023feature} incorporates perceptual and adversarial loss into their approach, utilizing a feature reuse network to overcome the limitations associated with relying solely on reconstruction loss. This adaptation helps prevent the generation of overly smooth SR images. Furthermore, they introduce dynamic filters and a spatial attention mechanism to mitigate the issues of irrelevant texture transfer and misalignment between the $LR$ and $REF$ images.
\citep{shim2020ssen} employs deformable convolutions to achieve adaptive alignment between $LR$ and $REF$ frames, effectively learning this alignment implicitly.


Mobile phone manufacturers have introduced a new dual-camera setup, featuring multiple lenses such as telephoto and wide-angle cameras. These cameras capture scenes from varying angles, enhancing the overall user experience.~\citep{wang2021dual, lee2022reference} leverage telephoto and wide-angle frames from mobile phones and address content disparities between the frames by transferring missing texture information from the wide-angle frames. However, this sometimes leads to blurriness at the edges of the resulting images. In contrast,~\citep{zou2023geometry} proposes an alternative approach. They employ a geometric matching method in combination with patch-matching to achieve more precise alignment between the $LR$ and $REF$ frames, enhancing the subsequent texture transfer process.


The application of attention-based networks in RefSR is quite natural, as it can facilitate patch matching through self-attention mechanisms. In many instances, attention-based RefSR approaches have focused on conducting global patch matching using a combination of hard and soft attention mechanisms~\citep{yang2020learning, lin2021residual, liu2022dsma}. Hard attention is typically employed to identify the most similar patches between the $REF$ and $LR$ images. This involves an exhaustive iteration over all possible patch pairs to determine the best matches. Subsequently, soft attention comes into play, assigning weights to each patch matching in the final result. The purpose of soft attention is to ensure that the contribution of each match is fairly considered and eliminating the risk of treating all patch pairs as equally significant.~\citep{jiang2021robust} integrates attention mechanisms into the fusion process. An attention map is generated based on extracted features and patch correspondences, with the goal of achieving a focused and enhanced fusion. \citep{cao2022datsr} integrates the Swin transformer to replace the traditional ResBlock for feature aggregation, significantly enhancing model performance, yet also increasing model complexity and runtime. While these attention-based strategies are valuable, they are often computationally demanding and they may not be sufficient to address all the complexities of RefSR for video sequences exhibiting complex motion.

\begin{figure*}[!t]
\centering
\includegraphics[width=\textwidth]{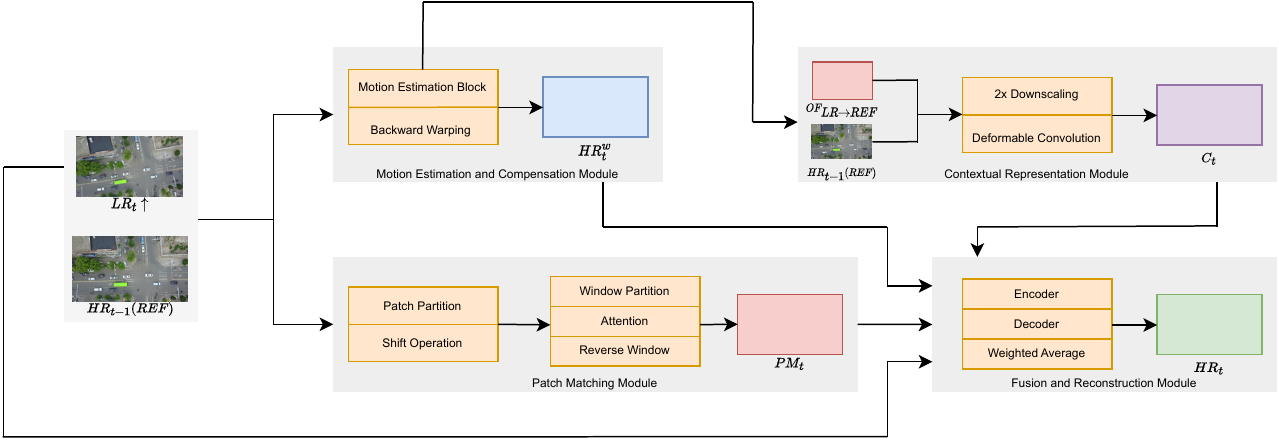}
\centering
\caption{Overall architecture of HSTR-Net. Motion estimation and compensation are applied to $LR$ and $REF$ frames to generate a warped reference frame. An attention-based patch-matching scheme is applied to input frames to make use of correspondences between similar textures. A contextual representation module extracts multi-scale features by applying deformable convolution at various resolutions. Lastly, a fusion and reconstruction module is used to generate the final $HR$ output frame.}
\label{fig:arch_diagram}
\end{figure*}

\subsection{Video Frame Interpolation (VFI)}

Video frame interpolation (VFI) is  employed to recover missing frames within a video sequence by leveraging information from neighboring frames. Typically, optical flow estimation between adjacent frames is utilized to create a warped reference frame, which is subsequently fused with additional features to generate the final output frame.
Recent work~\citep{lei2022flow, li2022h} proposes architectures that estimate optical flow between consecutive frames and utilizes deformable convolution for the motion compensation process. Deformable convolution offers enhanced adaptability to unseen data compared to traditional warping methods since it is trainable.~\citep{jin2022unified, lin2022mvfi,  li2022hybrid} predict bi-directional flows between consecutive frames, which are subsequently used for the warping process. They employ frame synthesis networks to refine the generated warped frames.~\citep{hu2022progressive} separates frame interpolation into two parallel processes, which are a temporal prediction network that focuses on motion-related tasks such as motion estimation and compensation, and a spatial prediction network that extracts visual content and features from neighboring frames.~\citep{plack2023frame} initially extracts features at various resolutions and uses them as inputs for each level of the network. Each level comprises two transformer blocks and a flow estimation block, and they utilize the output from the previous layer alongside the feature inputs.~\citep{zhou2023video} enhances motion estimation by gathering correlation information between frames and refining motion fields to generate more accurate motion information. This work also generates occlusion maps and synthesize the final frame using occlusion maps and estimated motions.~\citep{li2023amt} proposes extracting dense correlations before motion estimation to eliminate occlusions and blurs caused by large motions. This work generates multiple occlusion maps and motion fields to synthesize the final frame accurately, without occlusions or blurs.~\citep{liu2022ttvfi} employs two input frames and context features to synthesize an intermediate frame by estimating trajectories, reducing the impact of blur and distortion resulting from optical flow-based warping.~\citep{zhang2022optical} performs space-time video super-resolution and uses two $LR$ frames for motion estimation, followed by forward and backward warping operations to produce five $HR$ frames.~\citep{luo2022bi} leverages the information from four $HR$ frames to combine the extraction of multi-level spatio-temporal dependencies with the task of motion estimation and compensation to generate the final output.~\citep{li2022enhanced} utilizes four high-resolution ($HR$) input frames. These frames are employed for extracting spatio-temporal features and creating a multi-scale contextual representation, which serves as input to a fusion module based on deformable convolution.

In the context of wide-area monitoring, where video frame rates are typically low, conventional VFI methods struggle to cope with the sudden and rapid motions often encountered in wide-angle video feeds. This limitation arises from their heavy reliance on assumptions of constant and linear motion. Therefore, RefSR is a more viable approach for spatio-temporal resolution enhancement in wide-angle videos, since it can employ the $LR$ frame information at the generated output when accurate motion estimation or  $REF$-$LR$ registration is not possible. In other words, RefSR combines the advantages of SISR and VFI to generate  high-quality artifact-free HSTR videos.

%% file: tex/30_method.tex
\newcommand\vimeoh{256}
\newcommand\vimeow{448}

\section{Methodology}
\label{Methodology}

We propose a lightweight reference-based frame interpolation technique, namely HSTR-Net, to generate high spatio-temporal video feed from HSR-LFR and LSR-HFR video footages recorded by a dual camera setup. The architecture is inspired by RIFE \citep{huang2022rife} and illustrated in Figure \ref{fig:arch_diagram}. The missing $HR_{t}$ frame is synthesized by using an $LR_{t}$ frame from the LSR-HFR video and a neighboring $HR_{t - 1}$ or $HR_{t + 1}$ frame from the HSR-LFR video as a reference. Motion estimation and compensation techniques are applied as the initial block for generating a warped reference frame. Contextual representation extraction phase produces features and information at various scales using estimated optical flow and warped frame. Attention-based patch-matching module performs correspondence estimation between patches of $REF$ and $LR$ in a local manner to exploit similarities occurring between frames in two video feeds. Lastly, a fusion module synthesizes $HR_t$ by weighting the features gathered from prior stages.  The HSTR-Net architecture generates a high spatial quality $HR$ output through the coupled motion compensation and accurate texture transfer from $REF$ to $LR$ frame while considering execution time and power constrains. The local patch-matching mechanism provides necessary features for reliable and accurate texture transfer, since frames of time-synchronized video feeds have similar spatial textures. Furthermore, the reduced model complexity of single reference-based HSTR-Net enables deployment onto an embedded system composed of dual cameras. HSTR-Net achieves 4.62 fps on Jetson Xavier board with an input resolution of $672 \times 380$, which is sufficient for wide-area monitoring on drones.

\subsection{Motion Estimation and Compensation Module}
\begin{figure}[htbp]
\centering
\begin{subfigure}{0.9\linewidth}
\includegraphics[width=\linewidth]{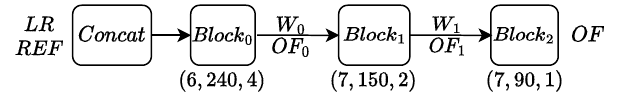}
\caption{Motion estimation module  (parameters (\textit{i,c,s}) given under each block)}\label{fig:motion_est1}
\end{subfigure}

\begin{subfigure}{0.95\linewidth}
\includegraphics[width=\linewidth]{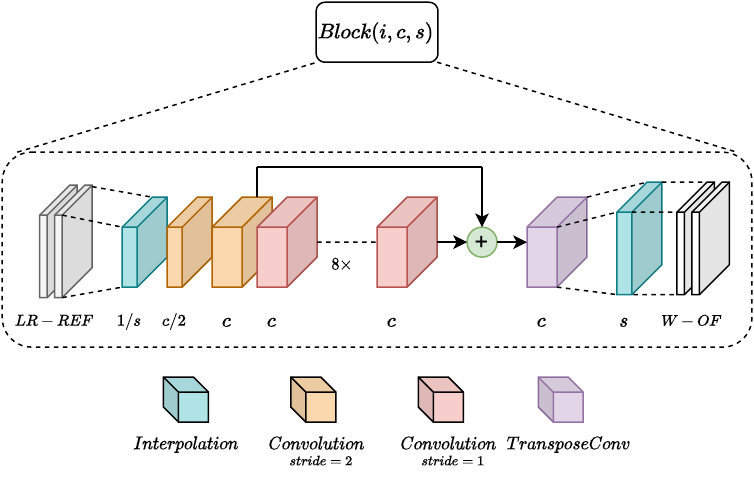}
\caption{Motion estimation block architecture (\textit{i, c, s} correspond to input channel size of a block, output channel size for convolution layers and scale factor, respectively)}\label{fig:motion_est2}
    
\end{subfigure}
\caption{Motion Estimation Module }
\label{fig:ifnet}
\end{figure}

HSTR-Net relies on accurate motion estimation and compensation since optical flow accuracy is important for the quality of the generated image in optical flow-based models. HSTR-Net motion estimation predicts optical flow from the $HR$ reference frame to the $LR$ frame, which is denoted as $OF_{LR \to REF}$. The motion estimation module consists of three consecutive blocks as shown in Figure \ref{fig:motion_est1} that extract motion at different scales; each block consists of downsampling and multiple convolutional layers as illustrated in Figure \ref{fig:motion_est2}. Each block has \textit{i, c, s} parameters that are input channel size, output channel size for convolution layers, and scale factor, respectively. Blocks 0, 1, and 2 have 6, 7, and 7 input channels and inputs are downsampled by scale factors of 4, 2, and 1, respectively for each block. $Block_0$ has 6 input channels and produces a warped reference image ($W_0$) with 3 channels and optical flow ($OF_0$) with 4 channels (representing both $OF_{LR \to REF}$ and $OF_{REF \to LR}$ with each having 2 channels) that are concatenated channel-wise and sent to the next block as input. $Block_1$ has 7 input channels and generates similar outputs as $Block_0$. $Block_2$ is the last block in the motion estimation module that has 7 input channels and calculates the final optical flows of $OF_{LR \to REF}$ and $OF_{REF \to LR}$.  Inside each block, the first two convolutional layers are used to match the output channel size and further downsample the input with a stride of 2. Then, 8 convolutional layers with the same input and output channel size are employed. Outputs of these two sub-blocks are added and fed to the transpose convolution layer, which has an output channel size of 4 and stride of 2. Interpolation is employed at the end to increase the resolution of the estimated optical flow field to that of the input frame.

\begin{figure*}[htbp]
\centering
\begin{subfigure}{0.9\linewidth}
\includegraphics[width=\linewidth]{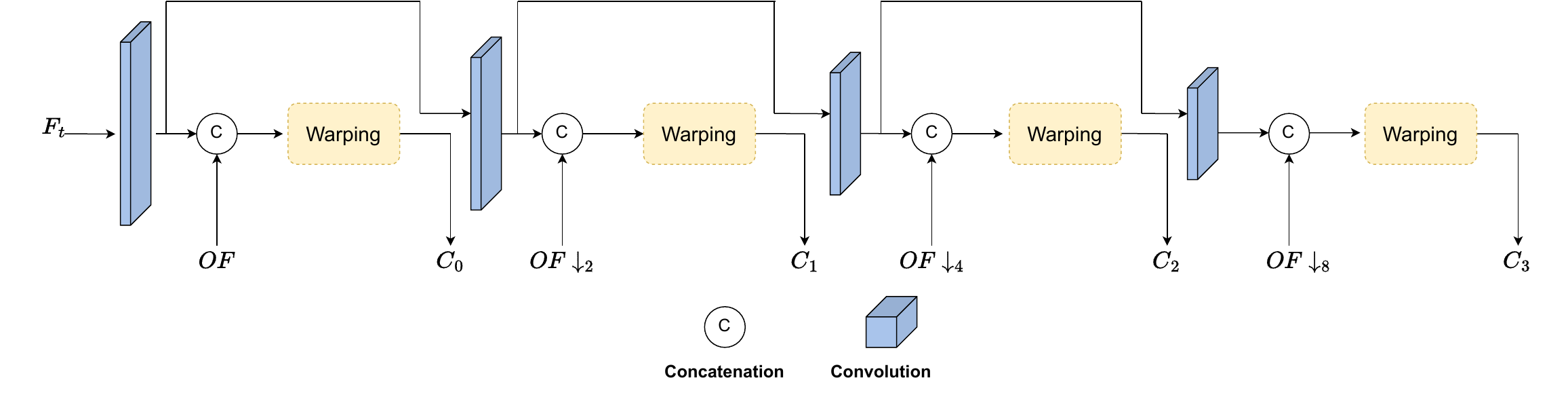}
\caption{$F_t$ and $OF$ denote input frame and optical flow respectively. In each level, the input feature map and $OF$ is downscaled by a factor of 2. $C_0$, $C_1$, $C_2$ and $C_3$ denote differently sized outputs of ContextNet.}\label{fig:warping_1}
\end{subfigure}
\begin{subfigure}{0.9\linewidth}
\includegraphics[width=\linewidth]{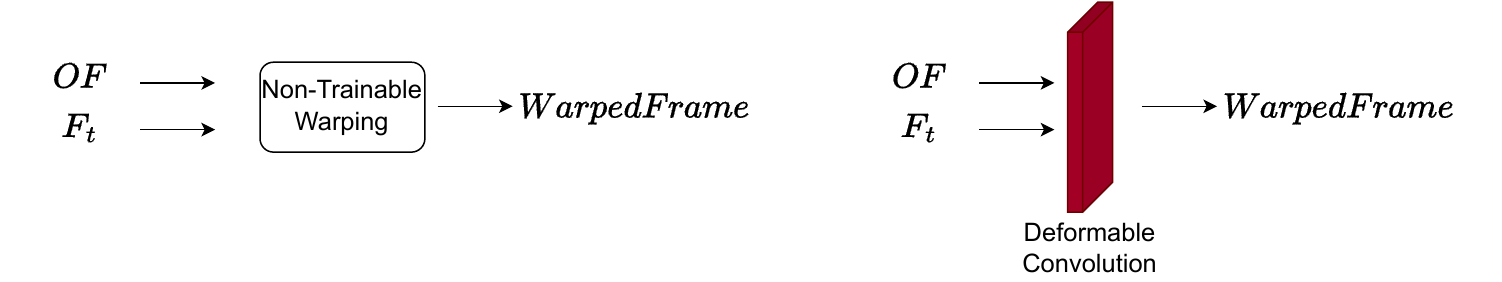}\caption{Left: Fixed warping approach, Right: Our adjustable deformable convolution approach}{\label{fig:warping_2}
    }
\end{subfigure}
\caption{Contextual Representation Module}
\label{fig:contextnet}
\end{figure*}

For motion compensation, a fixed warping operator is used to generate the warped $HR^W_t$ from the $REF$ frame. The warping module applies sub-pixel motion compensation to accurately obtain the target frame using vertical and horizontal flow components of $OF_{LR \to REF}$. $HR^W_t$ constitutes an initial prediction for $HR_t$; however it requires further refinement with the necessary features in order to obtain artifact-free high-quality results.

\subsection{Contextual Representation Module}

The contextual representation module is illustrated in Figure~\ref{fig:contextnet}. This module takes the $REF$ image and $OF_{LR \to REF}$ as input and generates feature maps $C_0, C_1, C_2, C_3$ at four different scales to support the subsequent fusion process.  The module gradually lowers the resolution of the input image by a factor of 2 using convolutional layers. Optical flow is also downsampled to match the feature map size. These convolutional layers have a stride of 2 and output channel sizes as (16, 32, 64, 128). Then, this module applies deformable convolution at each scale, treating optical flow as an offset in convolution for producing warped reference frame features at various resolutions. Deformable convolution is a trainable feature registration operator that provides more precise alignment for the features of the reference frame when compared to the traditional fixed warping. A learned warping operator can provide better alignment of extracted features with the synthesized frame. The deformable convolution module is a built-in \textit{PyTorch} function, which is efficient in terms of memory and computational complexity.

\subsection{Attention-Based Patch Matching Module}

In this section, we introduce our novel attention-based  architecture inspired by~\citep{liu2021swin}, which transforms self-attention for single input frame into a patch correspondence module for reference-based super-resolution. Texture transfer from $REF$ to $LR$ frame is an important phase that requires accurate detection of similar spatial textures. Prior work focuses on iterating over all possible pairs of patches in two images leading to inefficient models in terms of execution time and limiting the ability to deploy on power-constrained systems. However, our module is a  lightweight attention mechanism that performs patch-matching operations in various resolutions. In addition, the preceding methods rely on global patch-matching which is suitable for images with large texture differences. However, it is unable to exploit the local similarities between neighboring frames. Our patch-matching module is solely based on local correspondences with a window structure for detecting and transferring similar textures. 

\begin{figure*}[!t]
\centering
\includegraphics[width=\textwidth]{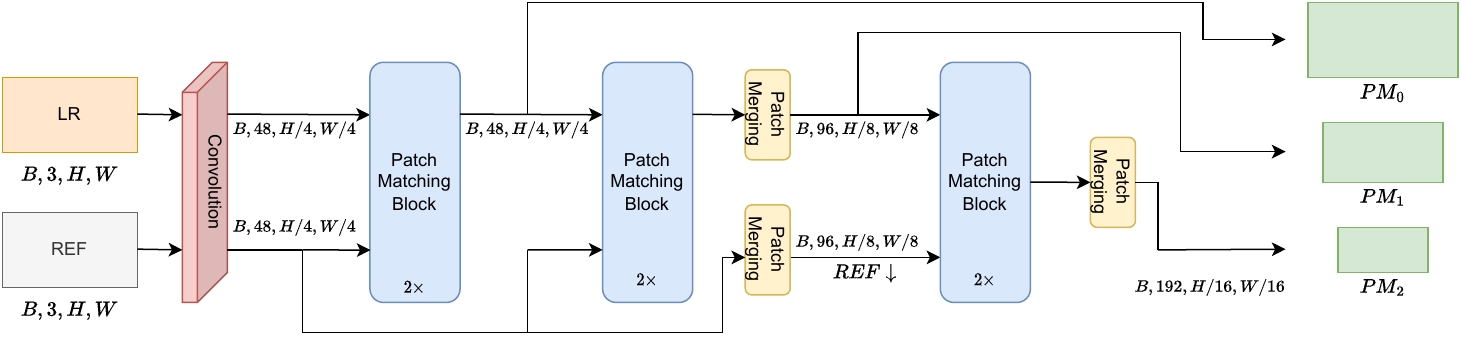}
\caption{High-level diagram of patch matching module. Given $LR$ and $REF$ frames, patch-partition is applied by a convolution layer. Later, patch-matching groups are applied to produce patch-matching outputs. Each output is used as an intermediate input for the next group and preserved for the final output. The last two groups apply patch-merging to decrease resolution by a factor of 2. }
\label{fig:swin_high_level}
\end{figure*}

In this module, $LR$ and $REF$ frames are initially partitioned into $4 \times 4$ patches by a convolution layer with a stride of 4 and output channel size of 48 and these patches are given as input to the sequential patch-matching blocks. Firstly, a shift operation is applied on these patches to extract features at different locations and scales,  and $3 \times 3$ windowed groups are formed from shifted patches. Then, a novel self-attention mechanism is applied on  these windows as detailed below. A pipeline of 6 patch matching blocks is created by forming 3 groups with each having 2 sub-blocks. Additionally, a patch merging mechanism is employed on the last two groups for increasing channel size and reducing the resolution of input feature maps. Consecutive patch matching groups have output channel sizes of $(48, 96, 192)$, respectively. This approach offers patch-matching in various resolutions and localizations that enable a better adjusted and robust fusion process. A high-level overview of this module is illustrated in Figure \ref{fig:swin_high_level}.

\textit{Patch Matching Block}:

This paper proposes a lightweight patch-matching block that performs self-attention on two given inputs. This block is the backbone of our attention mechanism and it applies attention at different scales by processing prior outputs of itself. 

The first  sub-block in the patch matching block combines patches into windows to calculate correspondence between patches within the same window. In the second sub-block, a shifting operation is applied on partitioned windows. The two successive sub-blocks are connected in order to introduce cross-patch correspondence and exploit the local information at various positions. This approach enables  focusing on the local similarity between neighboring patches and is a suitable method for video super-resolution, since similar textures lie in adjacent subareas of the neighboring patches. 
Then, the self-attention operation of~\citep{liu2021swin} is transformed into $LR$-$REF$ patch-matching as presented in Equation \ref{swin_eq}. The query ($Q$), key ($K$), and value ($V$) properties of both $LR$ and $REF$ frames are calculated for attention using linear layers. During attention calculation, two matrix multiplications are performed, which are $Q_{LR} \times K_{REF}$ and $Q_{REF}\times K_{LR}$, to obtain correspondences between $LR$ and $REF$ frames in both directions. The result of first matrix multiplication, $Q_{LR} \times K_{REF}$, is multiplied with $V_{REF}$ to obtain intermediate attention ${AT}_{i}$, since we are interested in transferring textures from $REF$. The other result, ${AT}_{LR}$, is multiplied with the ${AT}_i$ to incorporate the $LR$ attention information and get the final attention output of ${AT}$: 
\begin{equation}
\begin{aligned}
 &{AT}_{REF} = Q_{LR} \times K_{REF}\\
 &{AT}_{LR}  = Q_{REF} \times K_{LR} \\ 
 &{AT}       = {AT}_{LR} \times ({AT}_{REF} \times V_{REF})
\end{aligned}
\label{swin_eq}
\end{equation}

${AT}$ carries information on the patch-by-patch similarity measure between the two images, which provides a powerful reference for the texture transfer process. Unlike prior work, this approach is able to leverage the attention mechanism's advantages in an efficient manner resulting in faster and low-memory forward propagation.  Following the setting of \citep{liu2021swin}, a patch merging mechanism is used to create a hierarchical representation of feature maps at different resolutions. These feature maps are represented as $PM_0, PM_1, PM_2$ in Figure \ref{fig:swin_high_level} and they will be input to the fusion module to provide the matching high-frequency texture. Our patch-matching block is able to detect similar and relevant patches between $LR$ and $REF$ frames for the texture transfer process in an efficient way, leading to an optimized and high-performance network.


\subsection{Fusion Module}

\begin{figure*}[!t]
\centering
\includegraphics[width=\linewidth]{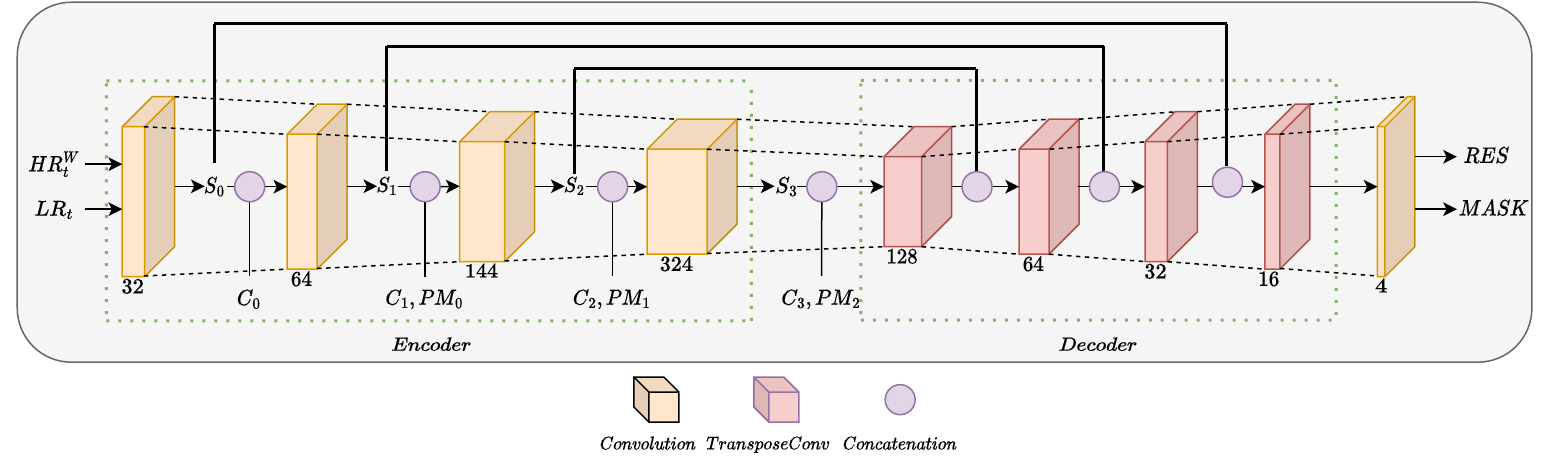}
\caption{Auto-encoder architecture of fusion module. The contextual features $C_i$ and patch correspondences $PM_{i-1}$ are concatenated with appropriate encoder layer outputs $S_i$ with matching resolutions and provided as input to the next encoder layer}
\label{fig:fusionnnet_diagram}
\end{figure*}

The fusion module is the final block to reconstruct the $HR_t$ frame. Our fusion block, $\mathcal{F}$, is a modified UNet\citep{ronneberger2015u} auto-encoder structure that processes contextual information in both finer and coarser resolutions, which is illustrated in Figure \ref{fig:fusionnnet_diagram}. In addition, it has residual connections between corresponding encoder and decoder layers, which exploit feature information from the earlier layers in the latter phases of the decoding process. The inputs of the auto-encoder block are $HR^W_t$ (warped reference frame), $LR_t$, $PM$ (patch correspondences), and $C$ (contextual representation block outputs). The outputs of the block are residual high-frequency  components  ($R$) and a weight mask $m$:
\begin{equation}
\begin{aligned}
R, m = \mathcal{F}(HR^W_t, LR, PM, C)\\
\end{aligned}
\label{fusion_eq}
\end{equation}

A subsequent reconstruction phase is employed to weigh the final contributions of $LR$ and the warped reference frame, $HR^W_t$, utilizing the residual high-frequency component $R$ and mask $m$ generated by auto-encoder architecture.  The weighted reconstruction is defined as below:
\begin{equation}
\begin{aligned}
 \hat{HR}_t = mHR^W_{t} + (1-m)LR + R
\end{aligned}
\label{fusion2_eq}
\end{equation}

The auto-encoder block has four downsampling and four upsampling layers with a final convolutional layer. Each downsampling layer consists of convolution followed by a ReLU operation where each upsampling layer is a combination of transpose convolution followed by a ReLU. All convolution and transpose convolution layers including the final layer have a kernel size of 3. Output channel sizes of consecutive downsampling and upsampling layers are (32, 64, 144, 304) and (128, 64, 32, 16) respectively. The last layer has 16 as the input channel size and 4 as the output channel size. Three channels of the output give residual components $R$ for each color band, and the other output is the weighting mask. Frame $HR_t$  is synthesized by using $R$ and mask $m$ as illustrated in Equation \ref{fusion2_eq}. The weighted average of the reference warped frame and $LR$ image is calculated using the mask, which balances the contribution of each input frame to the synthesized output. The residual $R$ compensates for any missing high-frequency texture that may not be matched with or transferred from the $REF$ frame.

%% file: tex/40_dataset.tex
\section{Experiments}
\label{sims}


\subsection{Datasets}
\label{sec:datasets}
HSTR-Net is trained and tested on $HR$-$LR$ image pairs from Vimeo90K\citep{xue2019video}, VisDrone\citep{VisDrone} and MAMI\citep{MAMI} datasets. Vimeo90K is a benchmark dataset for SR, VFI and RefSR applications. Vimeo90K septuplet dataset has 91,701 sequences arranged as 7 consecutive frames with a fixed resolution of $\vimeoh \times \vimeow$; HSTR-Net is trained and tested on the septuplet dataset. The triplet dataset of Vimeo90k has 73,171 3-frame sequences with resolution of $\vimeoh \times \vimeow$. Vimeo90K triplet dataset is only used for testing purposes to demonstrate the robustness of HSTR-Net across different content. Furthermore, VisDrone is an aerial imagery dataset consisting of drone-captured footages; this dataset is widely used in aerial object detection and tracking challenges. High-resolution VisDrone videos are downsampled to $672 \times 380$ resolution in order to simulate typical resolutions for wide-angle frame sequences. This dataset presents new  challenges because it includes fast drone motion/rotation, small moving objects, etc. Since both datasets only have $HR$ frames, whereas RefSR requires corresponding $LR$ frames,  4x bicubic downsampling followed by 4x bicubic upsampling is applied to synthetically generate the $LR$ frames. 

\begin{figure}[t]
\centering

\subfloat[$LR$ camera\label{fig:mami-1}]{%
  \includegraphics[width=0.23\textwidth,keepaspectratio]{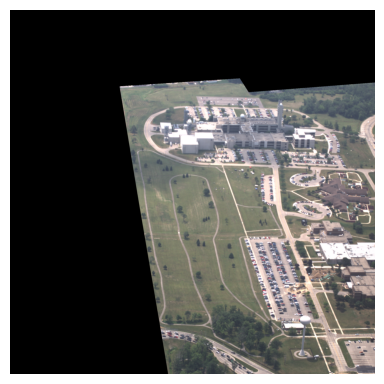}
}
\subfloat[$HR$ camera\label{fig:mami-2}]{%
  \includegraphics[width=0.23\textwidth,keepaspectratio]{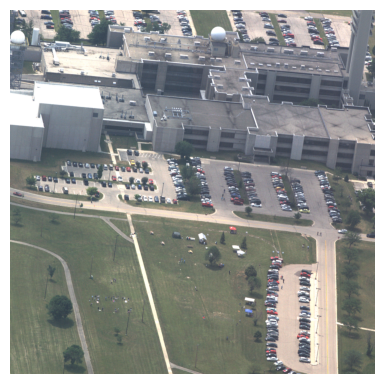}
}
\\
\subfloat[registered $HR$ image\label{fig:mami-3}]{%
  \includegraphics[width=0.23\textwidth,keepaspectratio]{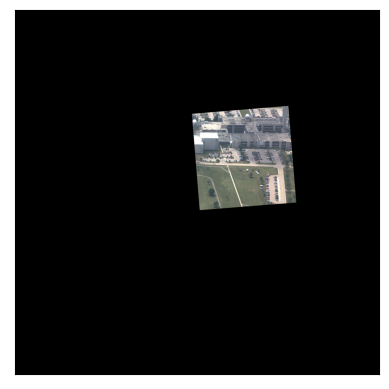}
}
\subfloat[registered $LR$ image\label{fig:mami-4}]{%
  \includegraphics[width=0.23\textwidth,keepaspectratio]{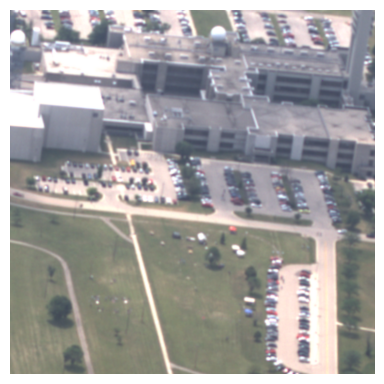}
}

\caption{Visuals from MAMI dataset}
\label{fig:mami-visual}
\end{figure}

In addition to these two datasets, a true dual-camera setup is employed using the Minor Area Motion Imagery (MAMI) \citep{MAMI} dataset. MAMI contains four $HR$ and one wide-angle $LR$ camera capturing the aerial video of a military campus. Figures \ref{fig:mami-1} and \ref{fig:mami-2}  show sample frames of the $LR$ and $HR$ cameras, respectively. $LR$ camera frames are matched and registered with $HR$ camera frames and cropped to $512 \times 512$ sized outputs in order to generate the MAMI dataset for RefSR purposes (see Figure \ref{fig:mami-4}). Our MAMI RefSR dataset consists of $8,018$ train and $1,983$ test samples with a resolution of $ 512 \times 512$. MAMI is a challenging dual-camera setup, since it includes camera misalignment, varying color and contrast, different orientations of the cameras, etc. Therefore, diverse augmentation techniques are necessary to train a generalized network, which is adapted to distortions and misalignments of the dual-camera setup. To the best of our knowledge, MAMI is the first real-life dual-camera dataset that specifically targets reference-based super-resolution in WAS. 

\subsection{Training and Experimental Setup}
HSTR-Net is trained on the Vimeo90K septuplet training set by using $HR_{t - 1}$ or $HR_{t + 1}$ frame as $REF$ for $LR_{t}$ with \%50 probability. Random cropping, rotation  and flipping are applied for data augmentation during training. Each frame is cropped randomly to 128x128 resolution. Random rotation varying from -10 to 10 degrees and flipping with \%20  probability are applied as additional data augmentation techniques. $L_1$ loss is used for training to synthesize sharp and qualitatively better-looking frames while slightly losing quantitative accuracy.  HSTR-Net is trained for 62 epochs with a learning rate of $10^{-5}$.

Peak signal-to-noise ratio (PSNR) and structural similarity index measure (SSIM) are used as metrics for evaluation. We adapt the input set configuration from~\citep{cheng2020dual} and designate the fourth frame of the septuplet test sequence as the $LR$ frame and the fifth frame as $REF$ for the evaluation of RefSR models. For VFI models,  the third and fifth frames are used as input to synthesize the fourth frame.~\citep{luo2022bi, li2022enhanced} models already provide quantitative results for the fourth frame utilizing the first, third, fifth, and seventh frames for inference. On the triplet dataset, the second frame is downsampled and used as $LR$ and the third frame becomes $REF$ for RefSR models, while the first and third frames are provided as input to the VFI models. Lastly, $LR_{t}$  and $HR_{t + 1}$ frames are used on VisDrone test sequences for RefSR model evaluations.

%% file: tex/50_results.tex
\input{tables/triplet_table_tex}
\input{tables/VisDrone_table_tex}

\subsection{Results and Discussions}
\label{sec:result}

We first present quantitative comparison results of RefSR techniques on Vimeo90K and VisDrone datasets. We employ state-of-the-art RefSR methods to compare against HSTR-Net. VFI results are also provided for reference. 
In Table \ref{tab:septuplet_table}, measurements regarding RefSR models are collected from septuplet dataset of our benchmark except~\citep{zheng2018crossnet}. The result for CrossNet~\citep{zheng2018crossnet} is taken from \citep{cheng2020dual}. VFI methods~\citep{luo2022bi, li2022enhanced}  provide PSNR/SSIM results on their respective papers;  ~\citep{huang2022rife} is re-evaluated on the  benchmark. In Table \ref{tab:triplet_table}, for the triplet dataset, the results of the tested methods are taken from their respective papers if they are available; otherwise the trained models are tested on the Vimeo90K triplet benchmark. In Table \ref{tab:VisDrone_table}, available RefSR models are compared on the VisDrone test set. To accurately assess the generalizability of each model to an aerial video dataset, we perform the simulations without any fine-tuning. In the tables, the best RefSR method is highlighted in bold, and the second best result is highlighted in red.

\input{tables/septuple_table_tex}

\begin{figure*}[t]
\centering

\begin{subfigure}{0.3\textwidth}
\includegraphics[width=\textwidth] {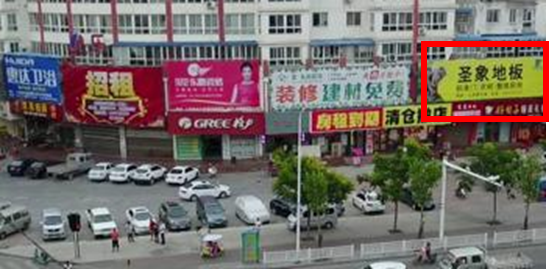}\caption*{Ground Truth}\label{fig:gt_vis}
\end{subfigure}
\begin{subfigure}{0.3\textwidth}
\includegraphics[width=\textwidth] {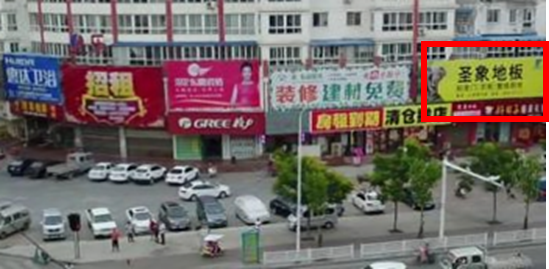}\caption*{HSTR-Net (30.25)}\label{fig:hstr_vis}
\end{subfigure}
\begin{subfigure}{0.3\textwidth}
\includegraphics[width=\textwidth] {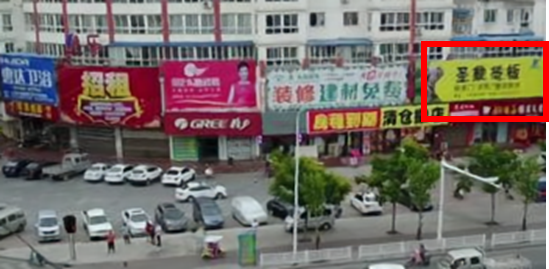}\caption*{MASA-SR (28.91)}\label{fig:masa_vis}
\end{subfigure}

\begin{subfigure}{0.3\textwidth}
\includegraphics[width=\textwidth] {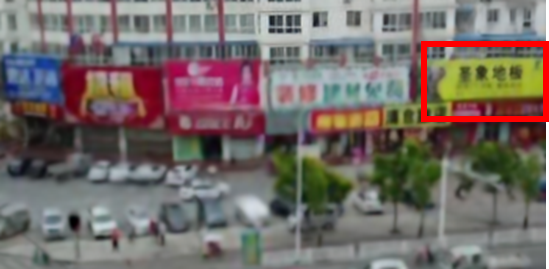}\caption*{C2-Matching (23.24)}\label{fig:c2_vis}
\end{subfigure}
\begin{subfigure}{0.3\textwidth}
\includegraphics[width=\textwidth] {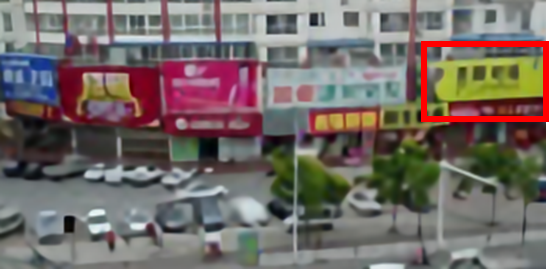}\caption*{TTSR (23.78)}\label{fig:ttsr_vis}
\end{subfigure}
\begin{subfigure}{0.3\textwidth}
\includegraphics[width=\textwidth] {figures/visuals/ttsr_vis.png}\caption*{AWNet (23.05)}\label{fig:awnet_vis}
\end{subfigure}

\begin{subfigure}{0.3\textwidth}
\includegraphics[width=\textwidth] {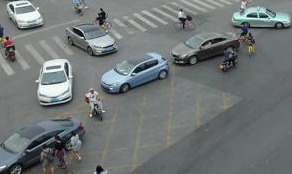}\caption*{Ground Truth}\label{fig:gt_sep}
\end{subfigure}
\begin{subfigure}{0.3\textwidth}
\includegraphics[width=\textwidth] {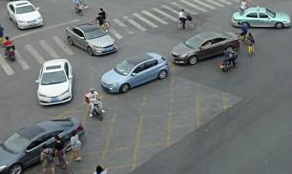}\caption*{HSTR-Net (33.30)}\label{fig:hstr_sep}
\end{subfigure}
\begin{subfigure}{0.3\textwidth}
\includegraphics[width=\textwidth] {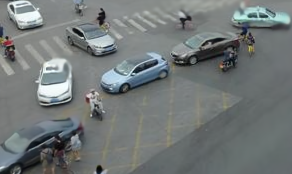}\caption*{MASA-SR (30.59)}\label{fig:masa_sep}
\end{subfigure}

\begin{subfigure}{0.3\textwidth}
\includegraphics[width=\textwidth] {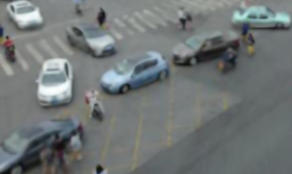}\caption*{C2-Matching (25.32)}\label{fig:c2_sep}
\end{subfigure}
\begin{subfigure}{0.3\textwidth}
\includegraphics[width=\textwidth] {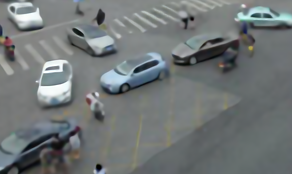}\caption*{TTSR (27.03)}\label{fig:ttsr_sep}
\end{subfigure}
\begin{subfigure}{0.3\textwidth}
\includegraphics[width=\textwidth] {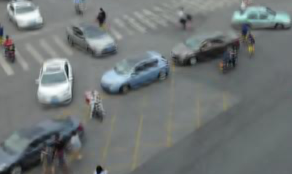}\caption*{AWNet (25.63)}\label{fig:awnet_sep}
\end{subfigure}

\caption{Qualitative comparison of different models on VisDrone dataset (PSNR in dB)}
\label{fig:visual}
\end{figure*}

As seen from Tables \ref{tab:septuplet_table} and \ref{tab:triplet_table}, HSTR-Net generates superior results when compared to state-of-the-art RefSR and VFI methods. HSTR-Net achieves $0.28$ dB and $0.09$ dB enhancement in PSNR metric with respect to second best RefSR result (i.e. AWNet) for septuplet and triplet datasets respectively, while reducing execution time by 47\% for Vimeo90K video resolution. Despite only using an $HR$ and an $LR$ frame, HSTR-Net surpasses models involving multi-frame setup such as~\citep{luo2022bi, li2022enhanced} as well as VFI models that use 2 $HR$ frames, since HSTR-Net's unique patch-matching module efficiently and accurately transfer high-frequency textures, which are obtained from $REF$ to $LR$ frame. Our unique patch-matching mechanism successfully contributes to the HSTR-Net, since it is designed specifically to capture local patch similarities between frames rather than global pixel or patch matching as in previous RefSR methodologies. Since adjacent patches are spatially similar for neighboring frames, the method provides accurate matching results at low complexity. Additionally, coupling motion estimation/compensation with the patch-matching method increases the model's robustness and accuracy by capturing dynamic movements between adjacent frames. 

The drone footages of VisDrone often include blurs, distortions caused by high-speed rotation and complex motion of the drone and moving objects at the scene. Therefore, accurate and reliable motion estimation and compensation complemented with necessary features and textures for the fusion process is essential for acquiring both quantitatively and visually pleasant RefSR results. HSTR-Net synthesizes better results compared to other RefSR models quantitatively on the VisDrone dataset, since it successfully combines motion compensation and texture transfer stages, as opposed to other tested models with limited capabilities for modeling complex motions. Additionally, local similarity embedding gives an edge to HSTR-Net, which is not present in global matching-based methodologies. For instance, AWNet achieves comparable performance to HSTR-Net on Vimeo90K; however the model fails to compete with HSTR-Net on the VisDrone dataset and gives more than 7 dB lower quality. HSTR-Net surpasses the second best approach (i.e. SRNTT) by 1.13 dB on the VisDrone dataset; however SRNTT is not suitable for real-time processing due to its high computational complexity. Despite being trained on the simple and low motion sequences of Vimeo90K, HSTR-Net model is capable of capturing the fast and complex motions of VisDrone sequences.
Table \ref{tab:VisDrone_table} also includes a model named HSTR-Net-\textit{f}, which is fine-tuned HSTR-Net on the VisDrone training dataset. Fine-tuning HSTR-Net increases PNSR by less than 1 dB, which shows that HSTR-Net is a well-generalized model. In fact, when the results from Tables \ref{tab:septuplet_table}, \ref{tab:triplet_table} and \ref{tab:VisDrone_table} are analyzed, it is clear that HSTR-Net is by far a more generalizable model than its competitors. 

Figure \ref{fig:visual} provides visual comparison between tested methods for selected frames of VisDrone dataset. As observed in Figure \ref{fig:visual}, HSTR-Net is able to capture fine details and generates visually sharp images. Other techniques are unable to cope with high rotations and increased speed of the drone leading to blurry and distorted outputs, since most existing architectures heavily rely on correspondence matching and do not include separate modules for motion estimation and compensation. Our approach is capable of synthesizing finer details, since it combines optical flow estimation and patch-matching schemes in an optimal way.

\begin{figure*}[t]
\centering

\begin{subfigure}{0.22\linewidth}
\includegraphics[width=\linewidth]{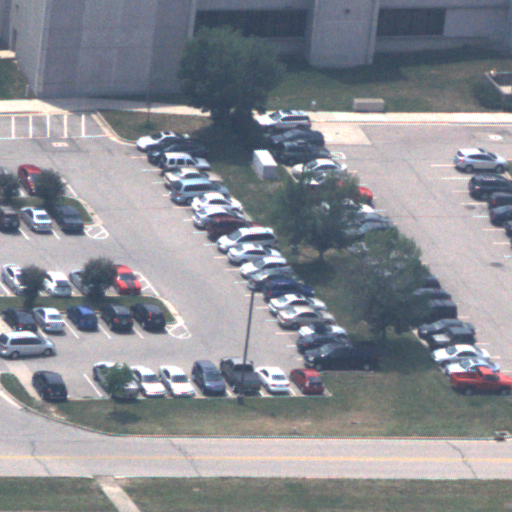}
\caption{Ground Truth}\label{fig:gt_mami}
\end{subfigure}
\begin{subfigure}{0.22\linewidth}
\includegraphics[width=\linewidth]{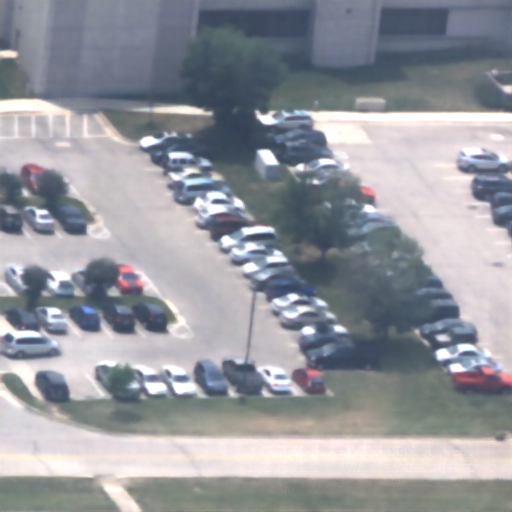}
\caption{HSTR-Net}\label{fig:hstr_mami}
\end{subfigure}
\begin{subfigure}{0.22\linewidth}
\includegraphics[width=\linewidth]{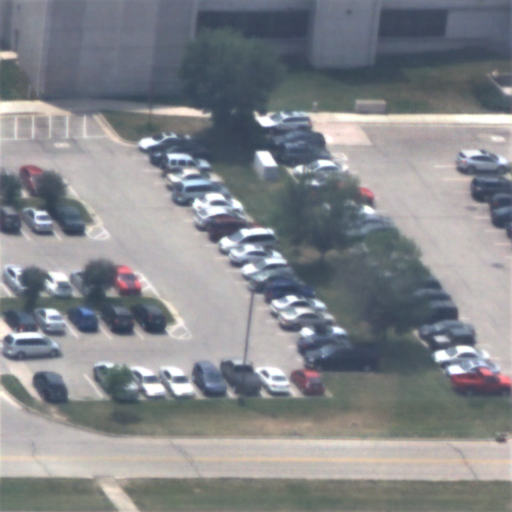}
\caption{FRFSR}\label{fig:frfsr_mami}
\end{subfigure}
\begin{subfigure}{0.22\linewidth}
\includegraphics[width=\linewidth]{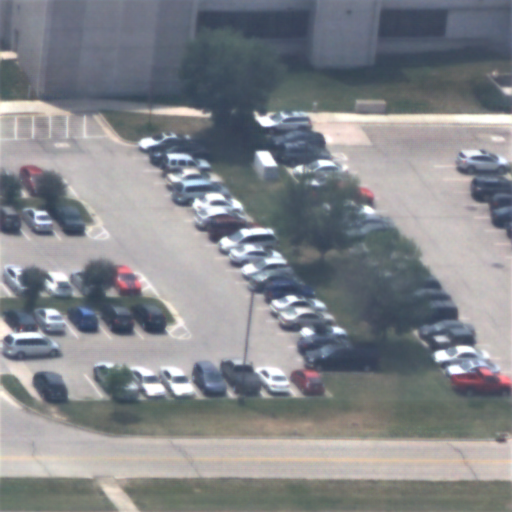}
\caption{DATSR}\label{fig:mami_datsr}
\end{subfigure}
\begin{subfigure}{0.22\linewidth}
\includegraphics[width=\linewidth]{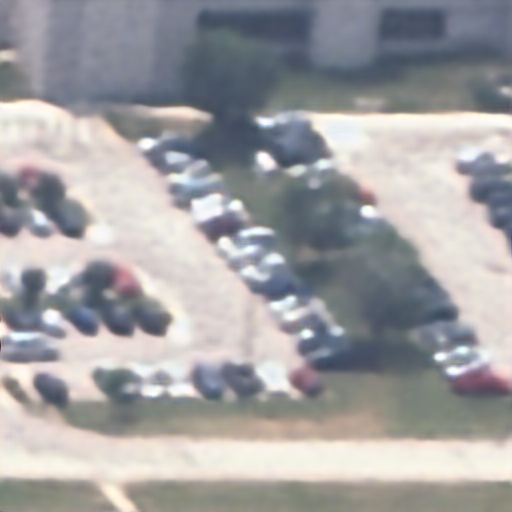}
\caption{MASA-SR}\label{fig:masa_mami}
\end{subfigure}
\begin{subfigure}{0.22\linewidth}
\includegraphics[width=\linewidth]{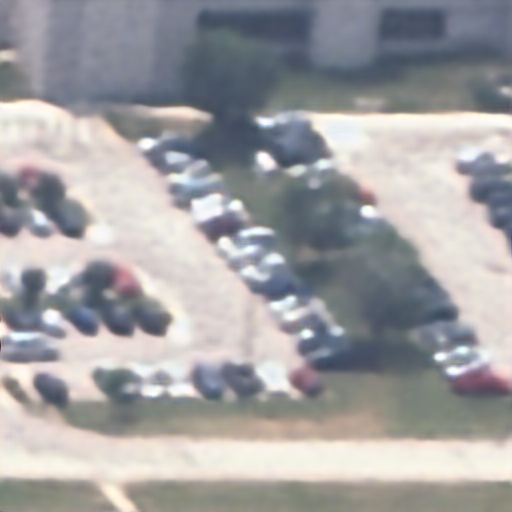}
\caption{SSEN}\label{fig:mami_ssen}
\end{subfigure}
\begin{subfigure}{0.22\linewidth}
\includegraphics[width=\linewidth]{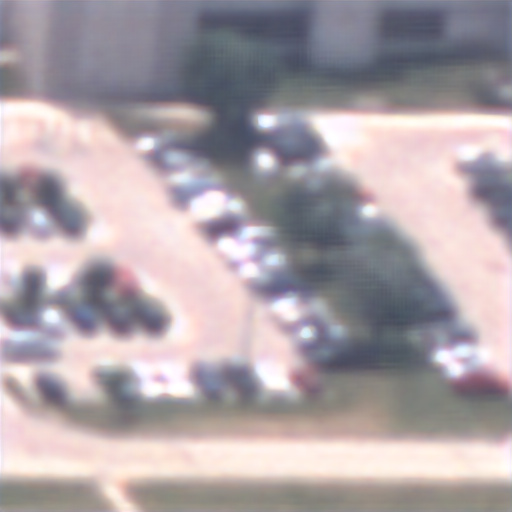}
\caption{TTSR}\label{fig:ttsr_mami}
\end{subfigure}

\caption{Qualitative comparison of different models on MAMI dataset}
\label{fig:mami_visual}
\end{figure*}

Next, HSTR-Net model is tested on the MAMI dataset to demonstrate that HSTR-Net is capable of adapting to real-life scenarios, thereby constitutes a suitable solution for WAS. For comparison, we fine-tune TTSR \citep{yang2020learning}, SSEN \citep{shim2020ssen}, FRFSR \citep{mei2023feature}, DATSR\citep{cao2022datsr}, and  MASA-SR \citep{lu2021masa} along with HSTR-Net on the MAMI training set. Since Vimeo90K is a synthetically generated training set, this fine-tuning is necessary in order for the models to adjust to the dual-camera issues, such as registration errors and differences in point spread function, color and contrast between the two camera frames. As observed in Table \ref{tab:mami-quant}, PSNR values are comparably lower due to the challenging characteristics of the MAMI dataset. However, HSTR-Net is superior to its counterparts by 1.11 to 4.76 dB, since the patch-matching scheme accurately complements the motion compensation module. Besides having higher PSNR and SSIM metrics, HSTR-Net generates visually sharp and pleasant images, thanks in part to the augmentation techniques employed during the training process to reduce the impact of distinct characteristics of $LR$ and $REF$ images. However, other methods are unable to compensate for the misalignments, contrast differences, etc., which eventually leads to blurry, color distorted and/or contrast distorted frames as presented in Figure \ref{fig:mami_visual}.

In Tables~\ref{tab:septuplet_table} and~\ref{tab:VisDrone_table} computational complexity of HSTR-Net is also compared against existing methods. Timing results in Table \ref{tab:septuplet_table} provide inference times for $\vimeoh \times \vimeow$-sized frames of Vimeo90K septuplet and triplet datasets. As seen from Table \ref{tab:septuplet_table}, HSTR-Net successfully balances accuracy and execution time, when compared to the second-best result provided by AWNet~\citep{cheng2020dual}. The efficiency of HSTR-Net is more observable when data resolution is increased. State-of-the-art RefSR methodologies cannot cope with the high resolution ($672 \times 380$) of the VisDrone dataset and fail to provide acceptable inference speeds for real-time applications. On the other hand, HSTR-net model complexity is able to linearly scale with the increase in resolution, which promises satisfactory frame rates for embedded deployment on a drone. HSTR-Net model achieves a $\%54$  speedup over second fastest RefSR model (i.e. AWNet) on the VisDrone dataset. We also deployed our model and two RefSR networks on Nvidia Jetson AGX Xavier Developer Kit~\citep{xavier} to showcase HSTR-Net's ability of low-overhead execution on an embedded platform. As presented in Table \ref{tab:xavier}, HSTR-Net outperforms other tested methods, outputting 4.62 fps on average, while MASA-SR and TTSR cannot even process a frame per second. This result indicates that HSTR-Net can be deployed in a drone setup involving dual cameras and can still achieve sufficient frame rates for detection, tracking or other surveillance tasks in wide-area monitoring.


\input{tables/mami_table_tex}
\input{tables/xavier_table_tex}

%% file: tables/triplet_table_tex.tex
\begin{table*}[!t]
\scriptsize
\caption{Quantitative comparison on Vimeo90k triplet dataset\label{tab:triplet_table}}
\centering
\begin{tabular}{c l c c c}
\hline
Method & Model & PSNR & SSIM\\
\hline

\multirow{9}{*}{\shortstack[c]{RefSR \\ 1LR + 1 HR}} 
                        &TTSR\citep{yang2020learning} & 31.61 & 0.927\\
                        &SSEN\citep{shim2020ssen} & 32.12  & 0.944\\
                        & MASA-SR\citep{lu2021masa} & 35.08 & 0.970\\
                        & SRNTT\citep{zhang2019image} & 35.14 & 0.972\\
                        & C2-Matching\citep{jiang2021robust} & 37.07 & 0.980\\
                        &FRFSR\citep{mei2023feature} & 37.43  & 0.981 \\
                        &DATSR\citep{cao2022datsr} & 37.96  & 0.980\\
                        &AWNet\citep{cheng2020dual} & \textcolor{red}{38.36} & \textbf{0.983}\\
                        & HSTR-Net & \textbf{38.45} & \textbf{0.983}\\

\hline
\hline
\multirow{7}{*}{\shortstack[c]{VFI \\ 2 HR}} & RIFE\citep{huang2022rife} & 35.61 & 0.978\\
                            & \citep{plack2023frame} & 36.08 & 0.979\\
                            & UPR-Net\citep{jin2022unified} & 36.42 & 0.981\\
                            & FGDCN\citep{lei2022flow} & 36.46 & 0.981\\
                            & AMT\citep{li2023amt} & 36.53 & 0.982\\
                            & TTVFI\citep{liu2022ttvfi} & 36.54 & 0.981\\
                            & DQBC\citep{zhou2023video} & 36.57 & 0.981\\
\hline

\multirow{1}{*}{\shortstack[c]{VFI \\ 4 HR}} & ESTF ~\citep{li2022enhanced} & 35.14 & 0.976\\ & & \\
\hline

\hline
\end{tabular}
\end{table*}

%% file: tables/visdrone_table_tex.tex
\begin{table*}[!t]
\scriptsize
\caption{Quantitative comparison on VisDrone dataset \label{tab:VisDrone_table}}
\centering
\begin{tabular}{c l c c r}
\hline
Method & Model & PSNR & SSIM & Time(ms)\\
\hline

\multirow{10}{*}{\shortstack[c]{RefSR \\ 1LR + 1 HR}} 
            &SSEN\citep{shim2020ssen} & 22.76  & 0.796 & 231.16\\
            & C2-Matching\citep{jiang2021robust} & 25.25 & 0.868 & 260.08\\
            & AWNet\citep{cheng2020dual} & 26.16 & 0.852 & \textcolor{red}{116.53} \\
            & TTSR\citep{yang2020learning} & 27.57 & 0.883 & 214.40\\
            & DATSR\citep{cao2022datsr} & 29.76  & 0.926 & 326.87\\
            & MASA-SR\citep{lu2021masa} & 30.65 & 0.940 & 160.30\\
            & FRFSR\citep{mei2023feature} & 31.12  & 0.953 & 248.32\\
            & SRNTT\citep{zhang2019image} & \textcolor{red}{32.17} & \textcolor{red}{0.959} & {3527.22}\\
            & HSTR-Net & \textbf{33.30} & \textbf{0.965} & \textbf{53.35} \\
\hline
            & HSTR-Net-\textit{f} & 34.13 & 0.969 & 53.35 \\


\hline

\hline
\end{tabular}
\end{table*}

%% file: tables/septuple_table_tex.tex
\begin{table*}[!t]
\scriptsize
\caption{Quantitative comparison on Vimeo90k septuplet dataset \label{tab:septuplet_table}}
\centering
\begin{tabular}{c l c c r}
\hline
Method & Model & PSNR & SSIM & Time(ms)\\
\hline

\multirow{10}{*}{\shortstack[c]{RefSR \\ 1LR + 1 HR}}
                            & TTSR\citep{yang2020learning} & 33.27 & 0.940 & 72.13\\
                            & SSEN\citep{shim2020ssen} & 34.45 & 0.958 & 84.32 \\
                            & SRNTT\citep{zhang2019image} & 35.76 & 0.975 & 1326.28\\
                            & MASA-SR\citep{lu2021masa} & 36.57 & 0.980 & 68.72\\
                            & C2-Matching\citep{jiang2021robust} & 38.40 & 0.983 & 90.82\\
                            & CrossNet~\citep{zheng2018crossnet} & 39.17 & 0.985 & -\\
                            & DATSR\citep{cao2022datsr} & 39.26 & 0.983 & 156.43\\
                            & FRFSR\citep{mei2023feature} & 39.55 & 0.985 & 123.65 \\
                            & AWNet\citep{cheng2020dual} & \textcolor{red}{39.66} & \textbf{0.986} & \textcolor{red}{56.02}\\
                            & HSTRNet & \textbf{39.94} & \textbf{0.986} & \textbf{29.56}\\

\hline
\hline
\multirow{1}{*}{\shortstack[c]{VFI \\ 2 HR}} 
& RIFE\citep{huang2022rife} & 35.78 & 0.972 & -\\ & & \\ 

\hline

\multirow{2}{*}{\shortstack[c]{VFI \\ 4 HR}} 
                            & BP3D~\citep{luo2022bi} & 35.76 & 0.959 & -\\
                            & ESTF ~\citep{li2022enhanced} & 35.75 & 0.961 & -\\
\hline

\hline
\end{tabular}
\end{table*}

%% file: tables/mami_table_tex.tex
\begin{table}[!t]
\scriptsize
\caption{Comparison of RefSR networks on MAMI dataset\label{tab:mami-quant}}
\centering
\begin{tabular}{l c c c}
\hline
Model & PSNR & SSIM\\
\hline
TTSR\citep{yang2020learning} & 20.58 & 0.678\\
SSEN\citep{shim2020ssen} & 22.32  & 0.719\\
MASA-SR\citep{lu2021masa} & 22.90  & 0.728\\
DATSR\citep{cao2022datsr} & 24.05  & 0.772\\
FRFSR\citep{mei2023feature} & \textcolor{red}{24.24}  & \textcolor{red}{0.785}\\
HSTRNet & \textbf{25.34}& \textbf{0.804}\\
\hline
\end{tabular}
\end{table}

%% file: tables/xavier_table_tex.tex
\begin{table}[!t]
\scriptsize
\caption{Execution times (in ms) for VisDrone dataset on Xavier platform. 
\label{tab:xavier}}
\centering
\begin{tabular}{c c c c}
\hline
 & HSTR-Net & MASA-SR & TTSR\\
\hline
\hline
Time (ms) & \textbf{216} & 1012 & 8955\\
\hline
\end{tabular}
\end{table}

%% file: tex/60_ablation.tex
\subsection{Ablation Study}

In this section,  several versions of HSTR-Net are compared to demonstrate the improvement in terms of  quantitative metrics with proposed modifications. Table \ref{tab:ablation} presents results for four different HSTR-Net models. Each model is separately trained on both Vimeo and VisDrone training sets. HSTRNet-\textit{b} is the baseline model performing motion estimation/compensation using IFNet from RIFE\citep{huang2022rife},   followed by contextual representation extraction and fusion processes. HSTRNet-\textit{i} is the model in which the motion estimation/compensation module is modified for RefSR purposes. Since single RefSR requires only the optical flow $OF_{LR \to REF}$, the original architecture is modified to produce a single warped reference frame, which in turn provides a simplified module with lower channel sizes and also with more accurate motion estimation. The model HSTRNet-\textit{d} incorporates deformable convolution, which uses a memory-efficient built-in PyTorch module, rather than non-trainable warping for the contextual feature extraction process. HSTR-Net is the original model presented in the paper that includes the unique patch-matching scheme as well. 

As observed in Table \ref{tab:ablation}, HSTRNet-\textit{b} model with general purpose optical flow estimation is limited by its ability to accurately match similar spatial textures between $REF$ and $LR$ frames, due to solely relying on  motion compensation. Therefore, designing a task-specific motion estimation and compensation module alongside a powerful yet efficient deformable convolution stage enhances the performance of the architecture. Additionally, video frames often include similar spatial content between neighboring frames  that can be exploited for better transfer of high-frequency features from $REF$ to $LR$ frame. Our Swin-based attention and patch-matching module proposes a new way of local patch-matching between the two frames in a memory-efficient manner, increasing the PSNR and SSIM metrics on both Vimeo septuplet and VisDrone datasets. 

HSTR-Net model is also applicable for increasing the frame rate by a factor more than 2. For 4x frame rate increase, HSTR-Net can be used to generate multiple $HR$ frames, given $LR_{\tau}$, $\tau = {t - 1, t, t + 1}$ and $REF_{\tau}$, $\tau = {t - 2, t + 2}$. Initially the model predicts $HR_{t - 1}$ and $HR_{t + 1}$. Then, $HR_{t}$ is predicted by using $(\hat{HR}_{t - 1}, LR_{t})$ and $(\hat{HR}_{t + 1}, LR_{t})$ frame pairs as input to HSTR-Net, for incorporating high-frequency features from both directions and increasing the accuracy. Lastly, the average of the two predictions is taken to synthesize the final $\hat{HR}_{t}$ prediction. The process is shown in Equation \ref{eq:intermediate}:

\begin{equation}
\begin{aligned}
 \hat{HR}_{t - 1} & =  \mathcal{M}(LR_{t - 1}, REF_{t - 2}) \\
 \hat{HR}_{t + 1} & =  \mathcal{M}(LR_{t + 1}, REF_{t + 2}) \\ 
 \hat{HR}_{t}^1   & =  \mathcal{M}(LR_{t}, \hat{HR}_{t - 1}) \\
 \hat{HR}_{t}^2   & =  \mathcal{M}(LR_{t}, \hat{HR}_{t + 1}) \\
 \hat{HR}_{t}     & =  0.5\hat{HR}_{t}^1   + 0.5\hat{HR}_{t}^2 \\
\end{aligned}
\label{eq:intermediate}
\end{equation}
where $\mathcal{M}$ denotes HSTR-Net. For the evaluation of 4x frame rate upsampling, the second and fifth $HR$ frames of the septuplet dataset are used for generating the third and fifth $HR$ frames. Then, the fourth frame is generated using these two intermediate reconstructed frames. Table \ref{tab:intermediate}  shows that competitive and highly accurate results are achieved even though intermediate $REF$ frames are synthesized by the model. On the VisDrone dataset, a consecutive group of five frames are used to simulate 4x upsampling. Results in Table \ref{tab:intermediate} show that the middle frame of the five-frame tuple yields average PSNR/SSIM results that are close to the other two generated frames, despite using lower quality $REF$ frames at the input. 

\input{tables/ablation_table_tex}
\input{tables/intermediate_table_tex}
\input{tables/interp_table_tex}

Table \ref{tab:hstr-interp} makes a comparison with our previous work called HSTR-Net-interp~\citep{suluhan2022dual}. HSTR-Net-interp utilizes $HR_\tau$, $(\tau=t-1,t+1)$ and $LR_\tau$, $(\tau=t-1, t, t+1)$ frames to generate $HR_{t}$ frame. This model has a similar architecture as the original HSTR-Net involving motion compensation, contextual representation extraction, and fusion process performed sequentially, but without any patch matching module. The model synthesizes warped reference frames, $HR^W_{\tau}$ $({\tau} = t - 1, t + 1)$ and $LR^W_{\tau}$ $({\tau} = t - 1, t + 1)$, using both $HR$ and $LR$ reference images during motion compensation process. Then, contextual representations, $(HR^C_{\tau}, LR^C_{\tau}, {\tau} = t - 1, t + 1)$ are calculated by gradually decreasing the input resolution for extracting valuable information and textures to aid the latter fusion process. Lastly, the fusion process takes $(HR^W_{\tau}, LR^W_{\tau}, HR^C_{\tau}, LR^C_{\tau}, LR_t)$ tuple as input, generating a mask and high-frequency residual channel at the output. Considering all these,  HSTR-Net-interp is bulky due to the use of 5 input frames, even though it has a simpler architecture when compared to HSTR-Net. As seen from Table \ref{tab:hstr-interp}, HSTR-Net-interp offers better results in terms of both PSNR and SSIM metrics. However, HSTR-Net is a better generalized model that can be applied at different frame rates and for different content with complex motion. HSTR-Net-interp requires fine-tuning when the motion complexity and the frame rate of the tested scenario changes, thereby providing a less generalized solution. Also,  HSTR-Net provides highly scalable performance in terms of time complexity and memory usage.


%% file: tables/ablation_table_tex.tex
\begin{table}[!t]
\scriptsize
\caption{Comparison of different HSTR-Net models on Vimeo septuplet and VisDrone datasets\label{tab:ablation}}
\centering
\begin{tabular}{l c c}
\hline
Method & Vimeo & VisDrone\\
& PSNR-SSIM & PSNR-SSIM\\
\hline
HSTRNet-\textit{b} & 36.86 - 0.978 & 32.02 - 0.957\\
HSTRNet-\textit{i} & 38.56 - 0.982 & 33.42 - 0.965\\
HSTRNet-\textit{d} & 39.34 - 0.984 & 33.93 - 0.968\\
HSTR-Net & \textbf{39.94 - 0.986}  & \textbf{34.13 - 0.969}\\
\hline
\end{tabular}
\end{table}

%% file: tables/intermediate_table_tex.tex
\begin{table}[!t]
\scriptsize
\caption{Quantitative results of intermediate HSTR-Net on Vimeo and VisDrone datasets, where 2nd, 3rd and 4th represents the respective frames in a five-frame group in which frames are ordered from 1 to 5 
 \label{tab:intermediate}}
\centering
\begin{tabular}{c c c}
\hline
Frame & Vimeo & VisDrone\\
& PSNR-SSIM & PSNR-SSIM\\
\hline
2nd & 39.95 - 0.986 & 34.35 - 0.971\\
3rd & 39.23 - 0.984 & 33.80 - 0.968\\
4th & 39.94 - 0.986 & 34.32 - 0.971\\
Avg & 39.71 - 0.985 & 34.16 - 0.970 \\
\hline
\end{tabular}
\end{table}

%% file: tables/interp_table_tex.tex
\begin{table*}[t]
\scriptsize
\caption{Comparison of Different HSTRNet models \label{tab:hstr-interp}}
\centering
\begin{tabular}{l c c c}
\hline
Method & 1 HR + 1 LR & 2 HR + 3 LR\\
& PSNR - SSIM - Time & PSNR - SSIM - Time\\
\hline
Vimeo Triplet & 38.45 - 0.983 - 29.56 & 40.62 - 0.988 - 64.11 \\
Vimeo Septuplet & 39.94 - 0.986 - 29.56 & 41.42 - 0.988 - 64.11\\
VisDrone & 33.30 - 0.965 - 53.35 & 35.06 - 0.973 - 81.79\\
\hline
\end{tabular}
\end{table*}

%% file: tex/70_conclusion.tex
\section{Conclusion}
\label{conclusion}
This paper addresses the problem of generating high spatio-temporal resolution (HSTR) video from multiple lower-resolution cameras using computational imaging techniques. The proposed solution involves a dual camera system that simultaneously captures high spatial resolution low frame rate (HSLF) video and low spatial resolution high frame rate (LSHF) video for the same scene. A novel deep learning architecture is introduced to fuse the two video feeds and generate high spatio-temporal resolution (HSTR) video frames using reference-based super-resolution (RefSR). The proposed model incorporates optical flow estimation, channel-wise attention, and spatial attention mechanisms to capture complex dependencies and fine motion between frames. Through experiments, it is demonstrated that the proposed model outperforms existing reference-based SR techniques in terms of peak signal-to-noise ratio (PSNR) and structural similarity index (SSIM) metrics. Moreover, the model is suitable for deployment on drones equipped with dual cameras, making it a promising solution for aerial surveillance applications.

\section*{Acknowledgement}
This work is supported in part by TUBITAK Grant - Project No: 118E891